\documentclass[final]{cvpr}

\usepackage{times}
\usepackage{epsfig}
\usepackage{graphicx}
\usepackage{amsmath}
\usepackage{amssymb}
\usepackage{url}
\usepackage{graphicx}
\usepackage{booktabs}
\usepackage{colortbl}

\usepackage[pagebackref=true,breaklinks=true,colorlinks,bookmarks=false]{hyperref}

\def\cvprPaperID{2} 
\def\confYear{CVPR 2021}
\usepackage{graphicx}
\begin{document}

\title{Small Object Detection for Near Real-Time Egocentric Perception in a Manual Assembly Scenario}

\author{Hooman Tavakoli\textsuperscript{1}, Snehal Walunj\textsuperscript{1}, Parsha Pahlevannejad\textsuperscript{2},\\ Christiane Plociennik\textsuperscript{2}, and Martin Ruskowski\textsuperscript{2}\\
\textsuperscript{1}SmartFactory-KL, \textsuperscript{2}German Research Center for Artificial Intelligence (DFKI)\\
}

\maketitle

\begin{abstract}
Detecting small objects in video streams of head-worn augmented reality devices in near real-time is a huge challenge: training data is typically scarce, the input video stream can be of limited quality, and small objects are notoriously hard to detect. In industrial scenarios, however, it is often possible to leverage contextual knowledge for the detection of small objects. Furthermore, CAD data of objects are typically available and can be used to generate synthetic training data. We describe a near real-time small object detection pipeline for egocentric perception in a manual assembly scenario: We generate a training data set based on CAD data and realistic backgrounds in Unity. We then train a YOLOv4 model for a two-stage detection process: First, the context is recognized, then the small object of interest is detected. We evaluate our pipeline on the augmented reality device Microsoft Hololens 2.
\end{abstract}

\section{Introduction}
\begin{figure}[htbp]
 \centering
  \includegraphics[width=\columnwidth]{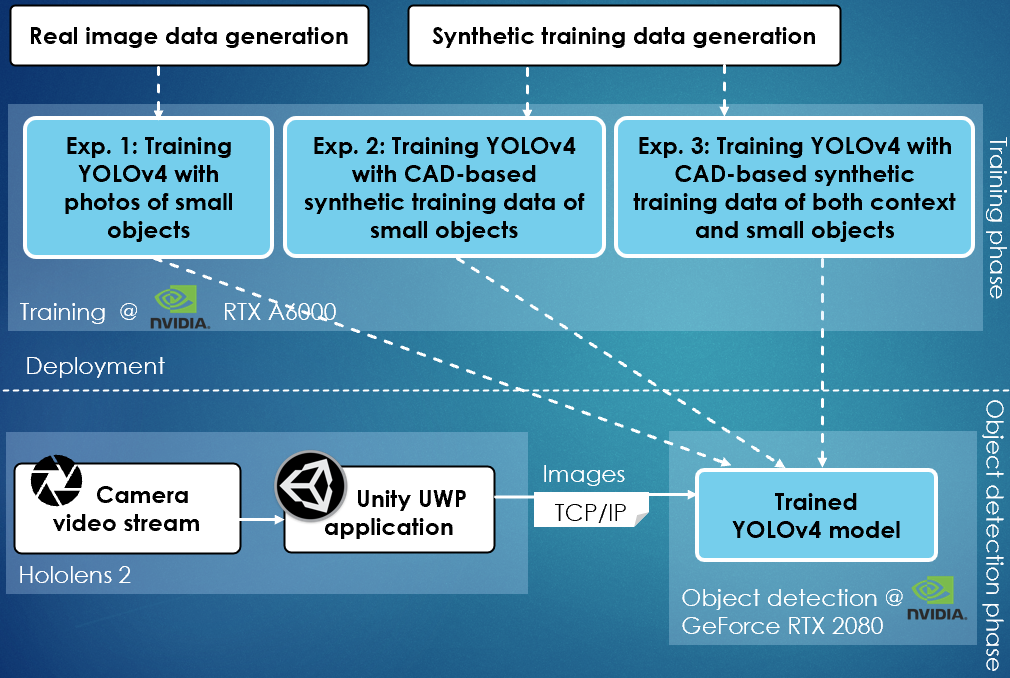}
  \caption{Experimental setup.}
  \label{pipeline}
 \end{figure}

\begin{figure*}[htbp]
  \centering
   \includegraphics[width=0.99\linewidth]{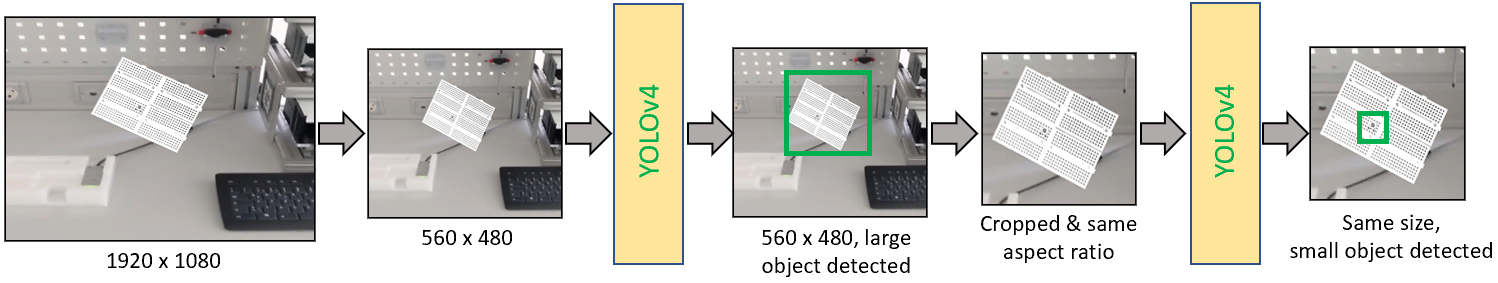}  
   \caption{Two-stage object detection pipeline.}
  \label{fig:experimentpipeline}
 \end{figure*}

For assisting workers in manual assembly, it is often required to detect objects manipulated by the worker via head-worn augmented reality devices \cite{su19}. This task of near real-time object detection in egocentric video streams is already a hard problem by itself: Viewpoints, angles, the distance to camera etc. change frequently, images are lower-quality due to poor focus and occlusion \cite{sab2019}, and there is typically a lack of high-quality training data. Moreover, object detection can usually not be performed on the augmented reality device due to resource constraints, and is typically offloaded to the edge or the cloud, which can introduce latency issues \cite{liu19}. The task becomes even more challenging when it comes to detecting small objects: These are typically represented by very few pixels in the input image, and the ground truth bounding box is small in comparison to the overall image, so the signal to be detected is small. Moreover, in the detection process, information is aggregated over the layers of the detector, hence the information representing small objects is gradually lost. Since hand-labelling images containing small objects is tedious, training data sets are typically of limited size. Furthermore, small objects frequently suffer from labelling errors \cite{Sola2020}. The problem of too little training data can be overcome by using synthetic training data. However, the problem of small signals remains. Fortunately, in manual assembly, it is often possible to leverage contextual knowledge. We propose a pipeline for near real-time object detection in egocentric video streams that first detects the context of a small object, then "zooms in" on the context and then detects the object itself (see Fig. ~\ref{fig:experimentpipeline}). We evaluate our approach in three experiments in a manual assembly scenario (see Fig. \ref{pipeline}).

\section{Related Work}

\textbf{Detection of Small Objects:} There is no common notion as to what is a small object. We follow the definition of Chen \etal who consider an object as small if the overlap area between the bounding box area and the image is 0.08\% to 0.58\% \cite{che2016}. Nguyen \etal evaluate state-of-the-art object detectors on small objects, finding that the YOLO series provides the best trade-off between accuracy and speed \cite{ngu2020}. Methods to improve model performance on small objects include: increasing image capture resolution \cite{Yang19}, increasing the model's input resolution \cite{Cou2020}, tiling the images \cite{Unel_2019_CVPR_Workshops}, data augmentation to generate more data \cite{kis2019}, auto learning model anchors \cite{Chen20}, and filtering out extraneous classes \cite{Sola2020}. Another possibility is to leverage contextual information to include visual cues that can help detect small objects \cite{xia2018}.\par
\textbf{Synthetic Training Data Generation:} Gajic \etal pursue a semi-synthetic data generation approach: foregrounds and backgrounds are captured separately from 3 positions and mixed \cite{gajic2020egocentric}. Elfeki \etal generate training images for egocentric perception based on 3rd person view (exocentric) using GANs \cite{elfeki2018person}. Cohen \etal's approach is similar to ours, albeit bigger objects are detected in a bus seat assembly process: CAD-based images are overlaid on industrial backgrounds, then 2D views with 162 camera viewpoints and 5 colors per viewpoint are generated \cite{cohen2020cad}.\par
\textbf{Egocentric Perception:} Sabater \etal evaluate YOLO on object detection in egocentric perception videos \cite{sab2019}. Farasin \etal describe a real-time object detection pipeline for the Microsoft Hololens, where object detection is carried out in the cloud \cite{fara20}. Fathi \etal leverage contextual knowledge in video image segmentation: They exploit the fact that there is a hand, an object and a background in each image \cite{fat11}.

\begin{figure*}[htbp]
 \centering
  \includegraphics[width=1\textwidth]{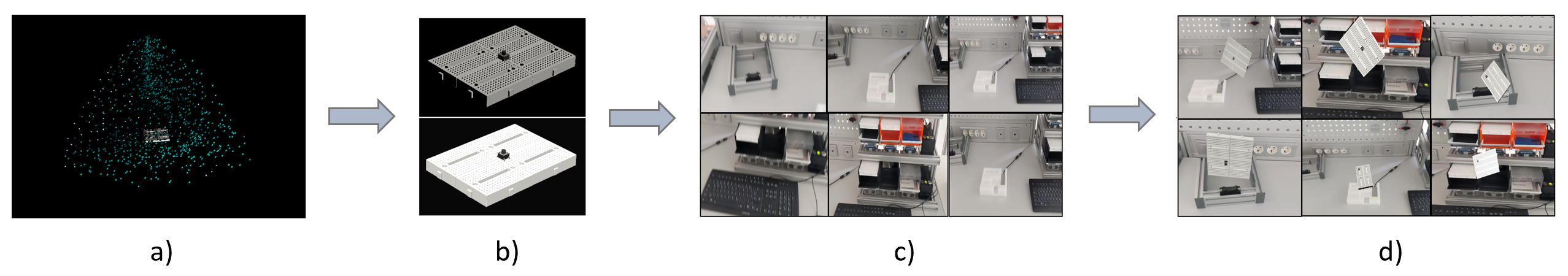}  
   \caption{Synthetic dataset generation, a) Camera positions around CAD object, b) Variation in illumination, c) Variation in backgrounds, d) Dataset generated from virtual scene.}
  \label{fig:synthpipeline}
 \end{figure*}

\section{Methodology}

We train on a dataset of synthetic images generated by combining real background images with objects generated from CAD models in Unity. This allows us to generate a comparatively large training data set with different viewpoints, angles, and illumination. The process of synthetic training data generation is described in Section \ref{sectionsynth}; the training process itself is described in Section \ref{sectionexp}. The trained YOLOv4 \cite{bochkovskiy2020yolov4} models were deployed on a laptop with an Nvidia GeForce RTX2080 as an edge device. For the actual detection, the camera stream of a Microsoft Hololens 2 is sent to the laptop, where object detection is performed. 
Our training and object detection pipeline is depicted in Fig. ~\ref{pipeline}.

\section{Synthetic Dataset}\label{sectionsynth}

Synthetic data generation is a much easier process and less time consuming than manually capturing and labelling a huge dataset of real images. Also, the CAD models in synthetic images can show better clarity than their real counterparts. Our synthetic images are created by loading 3D CAD models of the objects into a Unity 3D scene. In the 3D environment the virtual camera then captures images of the CAD object from various viewpoints around the object. We also considered the varying scale of the object by varying the distance of the camera from the object. This allowed us to simulate the process of capturing real image data \cite{cohen2020cad} within the renderer. In our use case, the objects have a relatively simple geometry and are perceived from a limited number of viewpoints. Hence, our viewpoints cover only a portion of the complete hemisphere around the object (see Fig. \ref{fig:synthpipeline}a). For generating the labels, we projected the bounding box around the CAD object onto the image screen and saved the bounding box coordinates in a label file.\par
\textbf{Domain Randomization:} For domain randomization we included various backgrounds captured from our workstation (see Fig. \ref{fig:synthpipeline}c). The real background images were used randomly on a plane in line with the camera axis, behind the CAD model. We also varied illumination and object viewpoints (see Fig. \ref{fig:synthpipeline}b). We used white lights and yellow lights with random illumination in the virtual scene while capturing images for the dataset. On the generated dataset, we carried out data augmentation in the form of random rotations of the labelled object and random shear in horizontal as well as vertical direction.\par
\textbf{Domain Adaptation:} YOLO is pre-trained on real images, so we adapted our dataset to the domain using photo realistic rendering on our CAD models. Since one of the object classes consists of 5mmx5mm size buttons, clarity was important in the training images. CAD objects with photo-realistic rendering achieved this clarity and thus allowed us to attain better quality of the small object images.

\section{Experiments and Results}\label{sectionexp}

We have defined 3 experiments to evaluate the results of our proposed methodology (see Fig. \ref{pipeline}): For Experiment 1, the YOLOv4 model has been trained with real images of small objects in front of realistic backgrounds. Experiment 1 serves as a baseline that enables us to compare the training performance of a small dataset of real images to that of a large dataset of synthetic images. For Experiments 2 and 3, the YOLOv4 model has been trained on synthetic training data (see Section \ref{sectionsynth}). All models have been trained on an Nvidia RTX A6000 GPU with CUDA 11.2. As we evaluated our approach on a manual assembly scenario, the small objects are tiny buttons, and the context is a breadboard where the buttons are to be assembled on.

\subsection{Experiment 1: Training with Conventional Photos}
For Experiment 1, we have trained our YOLOv4 model with a hand-labeled dataset of tiny buttons. We captured 95 photos using a Google Pixel 4a 5G camera. Then, with the RoboFlow platform \cite{Sola2020}, we applied preprocessing of resizing the images to 416x416 and augmentation of clockwise, counter-clockwise, upside-down and random shear horizontally and vertically. After applying the augmentations we had a dataset of 295 images.  In this experiment, we used the pretrained model \textit{yolov4.conv.137} and trained for 2000 epochs. For the test phase, we captured around 90 images (3904x2196px) from manual assembly, containing 221 tiny buttons, using the Hololens 2 camera with different lighting conditions, orientation, and distances from wearer. The results of Experiment 1 are shown in Table \ref{experiments_MAP}. They show that the model detected almost nothing; it could not generalise to the test data.

\begin{table*}[htbp]
\centering
\resizebox{0.75\textwidth}{!}{%
\setlength{\extrarowheight}{0pt}
\addtolength{\extrarowheight}{\aboverulesep}
\addtolength{\extrarowheight}{\belowrulesep}
\setlength{\aboverulesep}{0pt}
\setlength{\belowrulesep}{0pt}
\begin{tabular}{lllllllllllll} 
\toprule
                & \multicolumn{6}{l}{\textbf{mAP (full size)}}                                                                         & \multicolumn{6}{l}{\textbf{mAP (cropped)}}                                                                                                                                                                                               \\ 
\midrule
\textbf{IoU}    & \textbf{0.01}                        & \textbf{0.10} & \textbf{0.20} & \textbf{0.30} & \textbf{0.40} & \textbf{0.50} & \textbf{0.01}                        & \textbf{0.10}                        & \textbf{0.20}                        & \textbf{0.30}                        & \textbf{0.40}                        & \textbf{0.50}                         \\
\textbf{Exp. 1} & 0.3\%                                & 0\%           & 0\%           & 0\%           & 0\%           & 0\%           & 2.6\%                                & 0\%                                  & 0\%                                  & 0\%                                  & 0\%                                  & 0\%                                   \\
\textbf{Exp. 2} & {\cellcolor[rgb]{0.502,0.502,0.502}} & 44\%          & 26\%          & 4\%           & 0.6\%         & 0.03\%        & {\cellcolor[rgb]{0.502,0.502,0.502}} & {\cellcolor[rgb]{0.502,0.502,0.502}} & {\cellcolor[rgb]{0.502,0.502,0.502}} & {\cellcolor[rgb]{0.502,0.502,0.502}} & {\cellcolor[rgb]{0.502,0.502,0.502}} & {\cellcolor[rgb]{0.502,0.502,0.502}}  \\
\textbf{Exp. 3} & {\cellcolor[rgb]{0.502,0.502,0.502}} & 44\%          & 26\%          & 4\%           & 0.6\%         & 0.03\%        & {\cellcolor[rgb]{0.502,0.502,0.502}} & 70\%                                 & 69\%                                 & 58\%                                 & 27\%                                 & 8.5\%                                 \\
\bottomrule
\end{tabular}
}
\caption{This table illustrates the mAP for detection of the small button on the breadboard for our three experiments. The mAP is calculated according to \cite{Ralph20201}.}
\label{experiments_MAP}
\end{table*}

\subsection{Experiment 2: Training with Synthetic Images of Large and Small Objects}
For Experiment 2, we generated a training dataset of 1300 synthetic images of breadboards (large object) and 2500 synthetic images of tiny buttons (small object) in Unity (see Section \ref{sectionsynth}) and trained two separate models of YOLOv4 with these datasets separately. The training parameters were the same as in Experiment 1. For testing, we added 153 real images of breadboard -- partially occluded with tiny buttons and captured using Hololens 2 -- to the test dataset from Experiment 1. This enlarged test set was used in Experiments 2 and 3. The images were fed into the trained models for detecting the two separate classes of objects in parallel. The results are presented in Table \ref{experiments_MAP} (detection of button) and Table \ref{Breadboard_Results} (detection of breadboard). Table \ref{experiments_MAP} illustrates that the detection of buttons in Experiment 2 is far better than in Experiment 1: The larger dataset and the wider diversity of data assists the trained model to generalise to the new unseen test dataset.

\begin{table}[htbp]
\centering
\resizebox{0.45\columnwidth}{!}{%
\begin{tabular}{@{}llll@{}}
\toprule
\multicolumn{4}{l}{\textbf{mAP (breadboard)}}                                                                       \\ \midrule
\textbf{AP} & \textbf{AP50} & \textbf{AP75} & \textbf{mAP} \\
55\%        & 76\%          & 65\%          & 77\%         \\ \bottomrule
\end{tabular}
}
\caption{This table depicts the mAP \cite{Ralph20201} for the detection of the breadboard for Experiment 2 and 3 (IoU=0.50).}
\label{Breadboard_Results}
\end{table}

\subsection{Experiment 3: Two-Stage Process}
\begin{figure}[htbp]
 \centering
  \includegraphics[width=\columnwidth]{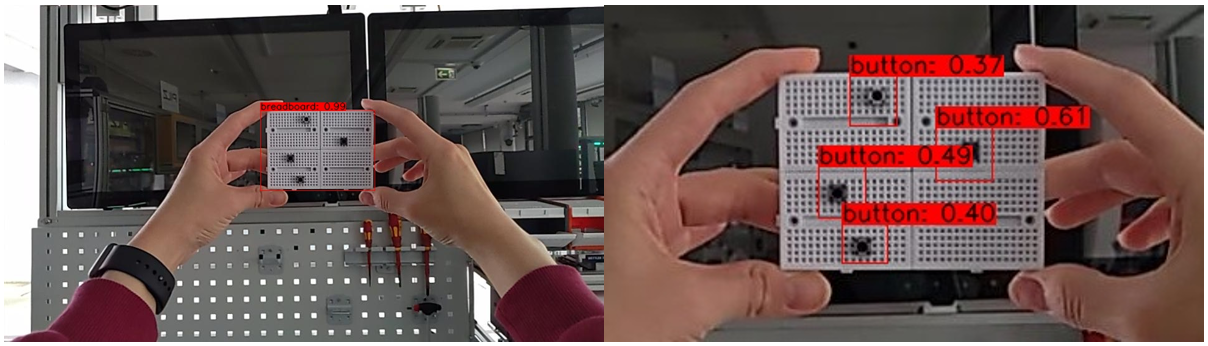}
  \caption{Two-stage object detection: The camera stream from the Hololens is analyzed on the edge device. First the context is detected (left), then the small objects (right).}
  \label{detection_hololens}
 \end{figure}

For Experiment 3, we used hierarchical detection and took into account the context (breadboard) of the small object (button): The first YOLOv4 network detects the breadboard in the test input image, we crop the part of the image containing the breadboard and feed it into the second YOLOv4 network (see Fig. \ref{fig:experimentpipeline}) which is trained to detect the small buttons. Both networks have the same configuration as in Experiment 2. Using this method, false detections of the breadboard (false negative or false positive) in the first stage directly affected the second stage, i.e. button detection, hence we included this point for Experiment 3 evaluation. The results are presented in Table \ref{experiments_MAP}: The mAP for the detection of small objects (cropped) in Experiment 3 is better than in Experiment 2. For IoU=10\% the mAP reaches 70\%. This means that hierarchical detection improves the detection of small objects. This pipeline runs at 9.4fps, which is considered near real-time (see Fig. \ref{detection_hololens}). 

\section{Conclusion and Future Work}
In this paper, we have shown that the task of detecting small objects in egocentric video streams in near real-time can be made easier, at least in industrial scenarios, if we leverage the specific circumstances of this domain: Since the context is usually clearly defined and well-structured, we can make use of it in the object detection process by first recognizing the context of a small object and then the object itself. Furthermore, CAD data of small objects is usually available and can be used to generate comparatively large amounts of high-quality training data. Our proposed two-stage small object detection pipeline performs significantly better than its single-stage counterpart. 
Future work includes displaying the object detection results as holograms on the Hololens, modelling objects and contexts in knowledge graphs for the object detection process, and leveraging eye-tracking data for small object detection.

\section{Acknowledgement}
We thank Daniel Pohlandt and Faranak Soleymani.

{\small
\bibliographystyle{ieee_fullname}
\bibliography{epic_2021}
}

\end{document}